\pdfoutput=1

\documentclass[11pt]{article}

\usepackage[]{EACL2023}
\usepackage{amsmath}

\usepackage{times}
\usepackage{latexsym}
\usepackage{graphicx}

\usepackage[T1]{fontenc}

\usepackage[utf8]{inputenc}

\usepackage{microtype}

\usepackage{inconsolata}

\title{BioUNER: A Benchmark Dataset for Clinical Urdu Named Entity Recognition}

\author{
Wazir Ali$^{\dagger}$, Adeeb Noor$^{\dagger\dagger}$, Sanaullah Mahar$^{\ddagger}$, Alia$^{\ddagger}$, Muhammad Mazhar Younas$^{\ddagger}$ \\
$^{\dagger}$Department of Data Science\\
Quaid-e-Awam University of Engineering, Sciences \& Technology, Pakistan \\
$^{\dagger\dagger}$Department of Information Technology\\
Faculty of Computing and Information Technology, King Abdulaziz University, Jeddah, Saudi Arabia \\
$^{\ddagger}$Department of Artificial Intelligence, Aror University, Sukkur, Pakistan
\\
Corresponding Author: \texttt{aliwazirjam@gmail.com}
}
\begin{document}
\maketitle
\begin{abstract}
In this article, we present a gold-standard benchmark dataset for Biomedical Urdu Named Entity Recognition (BioUNER), developed by crawling health-related articles from online Urdu news portals, medical prescriptions, and hospital health blogs and websites. After preprocessing, three native annotators with familiarity in the medical domain participated in the annotation process using the Doccano text annotation tool and annotated 153K tokens. Following annotation, the proposed BioiUNER dataset was evaluated both intrinsically and extrinsically. An inter-annotator agreement score of 0.78 was achieved, thereby validating the dataset as gold-standard quality. To demonstrate the utility and benchmarking capability of the dataset, we evaluated several machine learning and deep learning models, including Support Vector Machines (SVM), Long Short-Term Memory networks (LSTM), Multilingual BERT (mBERT), and XLM-RoBERTa. The gold-standard dataset BioUNER\footnote{The BioUNER dataset as well as the final model will be publicly available on Hugging Face.} serves as a reliable benchmark and a valuable addition to Urdu language processing resources.
\end{abstract}

\section{Introduction}
Clinical Named Entity Recognition (NER) is a specialized task in Natural Language Processing (NLP) that aims to identify and extract important entities from clinical narratives~\cite{averly2025entity}, such as diseases, symptoms, genes, and medications~\cite{luo2022biored}. These narratives typically include electronic health records (EHRs)~\cite{tutubalina2021russian}, clinical notes~\cite{li2021synthetic}, and patient records. Clinical NER plays~\cite{li2021synthetic} a critical role in real-world NLP applications by transforming unstructured medical text into structured information that supports clinical decision-making, medical research, and healthcare analytics. 

Benchmark or gold-standard datasets are key building blocks in NLP tasks, including clinical NER. Resource-rich languages such as English have abundant clinical NER datasets~\cite{huang2020biomedical}, ~\cite{luo2022biored}, as do Spanish~\cite{perez2021identifying}, German~\cite{liang2024building}, Chinese~\cite{guan2020cmeie}, French~\cite{bannour2024benchmark}, and Arabic~\cite{boudjellal2021abioner}.
In the case of Asian languages, general-domain NER datasets have been developed for Urdu~\cite{khana2016named}~\cite{kanwal2019urdu}, ~\cite{singh2012named}, Hindi~\cite{murthy2022hiner}, Bengali~\cite{haque2023b}, Sindhi~\cite{ali2020siner}, and other South Asian languages~\cite{goyal2008named}, ~\cite{ekbal2008language}. However, to the best of our knowledge, there is no publicly available dataset for clinical NER in Urdu.

To address the challenge of information extraction from clinical text, we collected real-world data from Urdu health blogs, medical articles, and clinical notes. The collected corpus was carefully cleaned and manually annotated to develop a Clinical Named Entity Recognition (NER) system for Urdu.

The main contributions of this article are as follows:
\begin{itemize}
\item We introduce BioUNER, a benchmark dataset for Urdu clinical NER, constructed from real-world data and annotated with gold-standard labels by expert human annotators.
\item We provide a comprehensive benchmark evaluation of the proposed BioUNER dataset using the original gold-standard annotations, including experiments on nested named entities.
\item We release the dataset publicly, along with masked language models fine-tuned for clinical NER in the Urdu language.
\end{itemize}

\section{Related Work}
Clinical NER primarily focuses on extracting Named Entities (NEs) such as genes, proteins, diseases, chemicals, and drugs from clinical records, and scientific literature. Progress in this area has been largely driven by benchmark datasets, particularly in English. An early resource, the GENIA corpus version 3.0 \cite{kim2003genia}, consists of 2K MEDLINE abstracts, comprising more than 400K words and nearly 100K annotations for biological terms. Subsequently, JNLPBA \cite{kimIntroductionBioentityRecognition2004} was developed based on the GENIA-3 corpus of MEDLINE abstracts and defined standard biomedical NEs types. This was followed by domain-focused datasets such as BC2GM \cite{smith2008overview}, NCBI-Disease \cite{dogan2012improved}, and the large-scale BioRED corpus \cite{luo2022biored}. Despite their impact, these resources differ in annotation schemes, NEs boundary definitions, and the treatment of nested NEs \cite{huang2020biomedical}.

Chinese clinical NER has been investigated at large scale, thus it also stands among rich resourced languages. The datasets such as CMeEE \cite{guan2020cmeie} and CMeEE-V2 \cite{zhang2025chinese}, and many others derived from electronic medical records. In terms of the development of NER System,  cross-lingual and joint learning approaches have also been explored to improve NE and relation extraction \cite{fu2024mmbert}, \cite{chen2020joint}. Moreover, PharmaCoNER \cite{ion2019racai} supports Spanish clinical NER, while Russian corpora such as RuDReC \cite{tutubalina2021russian} include drug and disease related annotations from user-generated texts.

For most South Asian languages, including Urdu, resources remain limited. The existing NER corpora and other language resources are primarily in the general domain. General datasets have been released for Urdu~\cite{khana2016named}~\cite{kanwal2019urdu}, \cite{singh2012named}, Hindi~\cite{murthy2022hiner}, Bengali~\cite{haque2023b}, Sindhi~\cite{ali2020siner}, and other regional languages~\cite{goyal2008named}, \cite{ekbal2008language}. However, these efforts focus on general NER rather than biomedical or clinical domains. To the best of our knowledge, no publicly available dataset currently exists for clinical NER in Urdu. Consequently, while languages such as English and Chinese benefit from established clinical benchmarks. 

\section{Data Collection \& Annotation}
This section presents the text collection methods, resources, preprocessing approach, annotation strategy, and, finally, the quality assessment of the proposed BioUNER dataset.

\subsection{Data Collection}
The clinical Urdu text used for the annotation of BioUNER was crawled from a diverse range of reliable sources, including clinical notes, doctors’ advice columns, health-related articles, and patient education resources. These sources include medical notes from Continental Hospitals\footnote{https://continentalhospitals.com/ur/blog/gut-bacteria-and-cancer-is-there-a-link/}, health-related articles from Daily Jang\footnote{https://jang.com.pk/category/latest-news/health-science}, as well as content from UrduPoint\footnote{https://www.urdupoint.com/health/}. Moreover, clinical notes were collected from The Aga Khan University Hospital Patient Education Materials\footnote{https://hospitals.aku.edu/pakistan/patients-families/Pages/patient\%20educational\%20material.aspx}, the Royal College of Psychiatrists (RCPsych)\footnote{https://www.rcpsych.ac.uk/mental-health/translations/urdu/physical-illness-and-mental-health}, and the Centre for Health Protection\footnote{https://www.rcpsych.ac.uk/mental-health/translations/urdu/physical-illness-and-mental-health}. The crawled Urdu text covers basic health awareness, mental health, pediatric guidance, and disease-related content, ensuring a rich and linguistically relevant clinical corpus for the annotation purpose.

\subsection{Preprocessing}
The preprocessing of newly crawled Urdu clinical text for annotation purposes involves several steps to ensure data quality and suitability. First, raw text collected from sources such as health articles, medical websites, and blogs was obtained, and HTML tags, non-Urdu characters, and noise (e.g., advertisements) were filtered out. Next, normalization addressed script-specific variations, including zero-width non-joiner handling and diacritic removal. Finally, sentence segmentation was applied to the cleaned corpus to prepare it for labeling using the Doccano annotation tool.

\subsection{Annotation}
\begin{figure*}[t]
    \centering
    \includegraphics[width=1\linewidth]{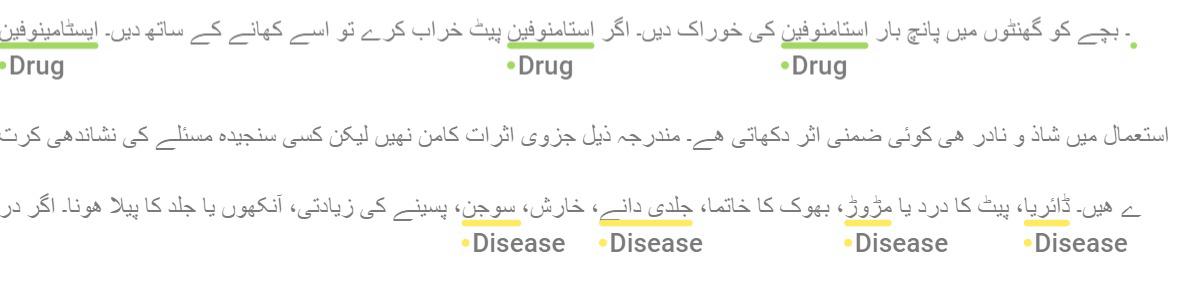}
    \caption{An example of annotation of BioUNER using Doccano text annotation tool}
    \label{fig:doccano}
\end{figure*}

We followed expert-defined guidelines and employed sequence labeling schemes such as BIES (Begin–Inside–End–Single)~\cite{zhang2022biomedical} to explicitly mark NE boundaries. Ambiguity~\cite{luo2022biored} is a major challenge in clinical NER annotation, including handling multi-word entities, resolving semantic ambiguities such as distinguishing genes from proteins, and addressing nested NEs. We used the Doccano text annotation tool~\cite{doccano} to annotate the BioUNER dataset (see Figure \ref{fig:doccano}) with six NE labels: CellLine, Chemical, Disease, Drug, Gene, and Protein. The BIES labeling scheme (see Table \ref{tab:sample_data}) was adopted to encode NE boundaries. 

\begin{table}[t]
    \centering
    \includegraphics[width=1\linewidth]{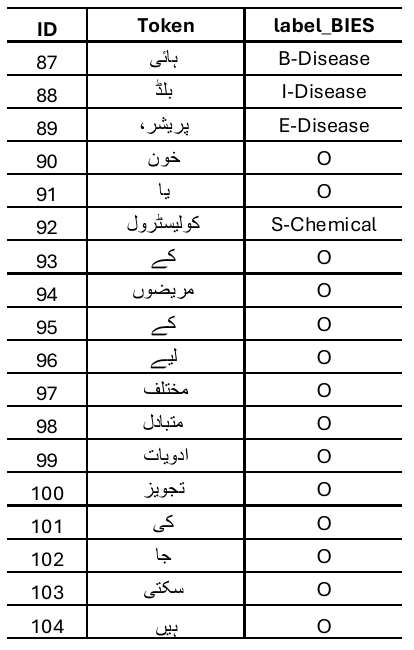}
    \caption{An example of annotated BioUNER dataset from token 87 to 104 }
    \label{tab:sample_data}
\end{table}

\subsection{Quality Assessment}
An inter-annotator agreement of $\kappa = 0.78$
 shows substantial agreement and points to a consistent annotation process. The BioUNER dataset was annotated by native-speaker annotators with domain knowledge, who followed the detailed annotation expert guidelines. Any disagreements were settled through adjudication to ensure the final dataset is reliable well as consistent.
 
 \subsection{BioUNER Dataset} 
 The newly proposed benchmark BioUNER dataset comprises six biomedical entity categories with varying frequency distributions. A total of 15,073 entity mentions are annotated across 153K words. The Disease category constitutes the highest number of NEs, indicating a strong emphasis on clinical conditions and diagnostic terms, followed by Chemical and Drug, reflecting substantial coverage. In contrast, molecular-level, Gene mentions, and CellLine NEs appear less frequently. In summary, the NEs distribution shown in Table \ref{tab:entity_counts} and BIES tag counts are shown in Table \ref{tab:bies_counts}. The distribution in Table \ref{tab:entity_counts} demonstrates a disease-centric corpus with balanced inclusion of pharmacological and molecular NEs, supporting its applicability for the clinical UNER task.
\begin{table}[h]
\centering
\caption{The complete entity type Counts in BioUNER dataset}
\begin{tabular}{lc}
\hline
\textbf{Entity Type} & \textbf{Count} \\ \hline
Disease   & 8959 \\ 
Chemical  & 2557 \\ 
Drug      & 2185 \\ 
Protein   & 604  \\ 
CellLine  & 520  \\ 
Gene      & 248  \\ \hline
\end{tabular}
\label{tab:entity_counts}
\end{table}

\begin{table}[h]
\centering
\caption{The overall statistics of BIES-Tag Counts in BioUNER dataset}
\begin{tabular}{lc}
\hline
\textbf{Tag} & \textbf{Count} \\ \hline
S-* & 5206 \\ 
B-* & 3888 \\ 
I-* & 2094 \\ 
E-* & 3883 \\ \hline
\end{tabular}
\label{tab:bies_counts}
\end{table}

\section{Experimental Setup}

\paragraph{Task Definition}
Given an input sentence 
$X = \{x_1, x_2, \dots, x_N\}$, 
where $x_i$ denotes the $i$-th word/token and $N$ represents the length of the sentence, the objective of Clinical NER is to assign a label to each token in the sentence.

Formally, the task is to learn a mapping from the input sequence $X$ to an output label sequence 
$Y = \{y_1, y_2, \dots, y_N\}$, 
where each label $y_i \in \mathcal{Y}$ and $\mathcal{Y}$ denotes a predefined set of NE labels such as \textit{Disease},  \textit{Gene}, and \textit{Protein}. 

In order to comprehensively evaluate the newly proposed BioUNER dataset, we exploit Conditional Random Fields (CRF)~\cite{quattoni2004conditional}, Hidden Markov Model (HMM), Support Vector Machines (SVM)~\cite{hearst1998support} are used as baseline models due to their effectiveness in sequence classification tasks. Furthermore, we employ Long Short-Term Memory (LSTM)~\cite{hochreiter1997long} network which capture contextual dependencies across sequences. In addition, we leverage pretrained multilingual transformer-based models including mBERT~\cite{pires2019multilingual} and XLM-RoBERTa~\cite{conneau2020unsupervised}, which provide rich contextualized representations and have demonstrated strong performance on low-resource and domain-specific NER tasks.

\section{Experimental Setup}

\subsection{Task Definition and Dataset}

In this study, we address Urdu biomedical NER as a supervised sequence labeling task. Given an input token sequence 
\(
X = (x_1, x_2, \dots, x_N)
\), 
the objective is to predict the corresponding label sequence 
\(
Y = (y_1, y_2, \dots, y_N)
\), 
where each \( y_i \) belongs to labe///     l set

\begin{equation}
\mathcal{Y} = \{\text{CellLine}, \text{Chemical}, \text{Disease}, \text{Drug}, \text{Gene}, \text{Protein}\}
\end{equation}
augmented with BIES boundary labels to encode NE spans.

For classical machine learning models, we adopt a 90/10 training-testing split. Neural and transformer-based models are trained using supervised fine-tuning on the training portion and evaluated on a held-out test set. All models are evaluated using Precision, Recall, and F1-score at the token level.

\subsection{Baseline Models}

To provide strong comparative baselines by implementing both traditional and neural architectures under consistent preprocessing and evaluation setup.

\subsubsection{Support Vector Machine (SVM)}
We exploit the SVM  as a token-level classifier. Each token \( x_i \) is represented using TF-IDF features extracted from character \( n \)-grams (\( n = 2 \) to \( 5 \)), which capture subword morphological patterns in BioUNER. This model serves as a non-contextual baseline, as it performs token-level classification without modeling label transitions.

\subsubsection{CRF}

To incorporate label transition dependencies, we implement a linear-chain CRF. Given a sequence \(X\), the conditional probability of a label sequence \(Y\) is modeled as:

\begin{equation}
P(Y|X) = 
\frac{1}{Z(X)}
\exp \left(
\sum_{i=1}^{N}
\left(
W^\top \phi(x_i, y_i)
+
T_{y_{i-1}, y_i}
\right)
\right)
\end{equation}

where \( \phi(x_i, y_i) \) denotes feature functions derived from the token representation, \( T_{y_{i-1}, y_i} \) represents learned transition scores between adjacent labels, and \( Z(X) \) is the partition function defined as:

\begin{equation}
Z(X) =
\sum_{Y'}
\exp \left(
\sum_{i=1}^{N}
\left(
W^\top \phi(x_i, y'_i)
+
T_{y'_{i-1}, y'_i}
\right)
\right)
\end{equation}

while the decoding is performed using:
\begin{equation}
\hat{Y} = \arg\max_Y P(Y|X)
\end{equation}
employed via  Viterbi decoding algorithm. In our case, CRF provides a structured prediction baseline that explicitly models entity boundary consistency under the BIES scheme.

\subsubsection{LSTM Model}
For capturing the contextual dependencies across tokens, we implement an LSTM-based architecture. Each token is mapped to a 100-dimensional embedding vector and processed sequentially through an LSTM layer with 128 hidden units:
\begin{equation}
h_i = \text{LSTM}(x_i, h_{i-1})
\end{equation}
Token-level label probabilities are computed as:
\begin{equation}
\hat{y}_i = \text{softmax}(W h_i + b)
\end{equation}  
In LSTM setting, we  apply dropout with a rate of 0.1 to deal with the overfitting, set maximum sequence length to 50 tokens, 10 epochs with a batch size of 32 using the Adam optimizer and categorical cross-entropy loss. 

\subsection{Transformer-Based Models}

To further strengthen performance in this low-resource biomedical setting, we fine-tune multilingual transformer encoders that provide contextualized token representations through self-attention.

\subsubsection{mBERT}
We fine-tune the mBERT, pretrained on more than 100 languages using masked language modeling. Given input sequence \( X \), contextual representations are obtained as:

\begin{equation}
H = \text{Transformer}(X)
\end{equation}

where,

\begin{equation}
H = (h_1, h_2, \dots, h_N)
\end{equation}

Token-level predictions are computed as:

\begin{equation}
\hat{y}_i = \text{softmax}(W h_i + b)
\end{equation}

Fine-tuning is performed for 3 epochs with a learning rate of \( 5 \times 10^{-5} \), batch size of 16, and weight decay of 0.01. Evaluation is conducted after each epoch.

\subsubsection{XLM-RoBERTa}
We further fine-tune XLM-RoBERTa-base, which is pretrained on a substantially larger multilingual corpus. Training is conducted for 5 epochs using a learning rate of \( 3 \times 10^{-5} \), batch size of 16, weight decay of 0.01, and maximum sequence length of 128 tokens. Logging is performed every 50 steps, with evaluation at the end of each epoch. Predictions are generated like mBERT.

\section{Results and Analysis}

\subsection{Baseline Performance}
The SVM baseline (see Table \ref{tab:baseline_results}) shows that lexical and surface-level features provide useful signals for Clinical NER. However, its  inability to effectively model contextual dependencies limits its performance, particularly for ambiguous entities and multi-token expressions. The LSTM improves the performance by capturing sequential dependencies and contextual information which leads to the boundary detection and overall classification performance compared to SVM and CRF respectively. 
\begin{table}[h]
\centering
\caption{Performance of baseline models evaluated using standard metrics—precision, recall, and F1-score. Boldfaced values indicate the best results.}
\begin{tabular}{lccc}
\hline
Model & Precision & Recall & F1-score \\
\hline
SVM  & 0.65 & 0.66 & 0.65 \\
CRF  & 0.93 & 0.93 & 0.93 \\
LSTM & \textbf{0.95} & \textbf{0.95} & \textbf{0.95} \\
\hline
\end{tabular}
\label{tab:baseline_results}
\end{table}

\subsection{Transformer-Based Performance}
The mBERT and XLM-RoBERTa transformer-based models outperform traditional baselines in contextual modeling. mBERT achieves competitive results through bidirectional contextual encoding across multiple languages. However, XLM-RoBERTa yields superior overall performance, particularly in recall and F1-score, indicating improved handling of ambiguous and multi-token biomedical NEs.
\begin{table}[h]
\centering
\caption{Performance of fine-tuned transformer models on the BioUNER dataset. Boldfaced values indicate the best results.}
\begin{tabular}{lccc}
\hline
Model & Precision & Recall & F1-score \\
\hline
mBERT & 0.72 & 0.71 & 0.72 \\
XLM-RoBERTa & \textbf{0.72} & \textbf{0.73} & \textbf{0.73} \\
\hline
\end{tabular}
\label{tab:transformer_results}
\end{table}

\section{Conclusion and Future Work}

In this study, we introduced the BioUNER gold standard benchmark dataset for clinical Urdu NER. The dataset was constructed by crawling clinical text from hospital and health blog websites, followed by preprocessing to ensure data quality and consistency. Annotation was performed using the Doccano text annotation tool with the involvement of three native annotators, achieving an inter-annotator agreement score of 0.78, which reflects substantial agreement and reliability of the annotated corpus. The evaluation results demonstrate that the LSTM model achieved the best overall performance with an F1-score of 0.95, outperforming the transformer-based models. Among the transformers, XLM-RoBERTa slightly surpassed mBERT, achieving an F1-score of 0.73, although both remained below the performance of the sequence-based neural baseline under the current experimental setup. In future, we plan to extend the size and diversity of the BioUNER dataset to improve model generalization and robustness. 

\section*{Acknowledgment}
The project was funded by KAU Endowment (WAQF) at king Abdulaziz University, Jeddah, Saudi Arabia. All the authors, therefore, acknowledge with thanks WAQF and the Deanship of Scientific Research (DSR) for technical and financial support.
\bibliography{anthology,custom}
\bibliographystyle{acl_natbib}

\end{document}